\documentclass[10pt,twocolumn,letterpaper]{article}

\usepackage{cvpr}
\usepackage{times}
\usepackage{epsfig}
\usepackage{graphicx}
\usepackage{amsmath}
\usepackage{amssymb}
\usepackage{tabularx}
\usepackage{multirow}
\usepackage{makecell}
\usepackage{booktabs}
\usepackage{multirow}
\usepackage{float}
\usepackage{cite}

\usepackage[breaklinks=true,bookmarks=false]{hyperref}

\cvprfinalcopy 

\def\cvprPaperID{8202} 

\ifcvprfinal\fi
\begin{document}
\newcommand{\modelnamewithoutspace}{Lip2Wav} 
\newcommand{\modelname}{Lip2Wav } 
\newcolumntype{Y}{>{\centering\arraybackslash}X}

\title{Learning Individual Speaking Styles for Accurate Lip to Speech Synthesis}

\author{K R Prajwal\thanks{Both authors have contributed equally to this work.}\\
IIIT, Hyderabad\\
\and
Rudrabha Mukhopadhyay\footnotemark[1]\\
IIIT, Hyderabad\\
\and
Vinay P. Namboodiri\\
IIT, Kanpur\\
\and
C V Jawahar\\
IIIT, Hyderabad\\
\and
{{\tt\small\{prajwal.k,radrabha.m\}@research.iiit.ac.in, vinaypn@iitk.ac.in, jawahar@iiit.ac.in}}
}

\maketitle

\begin{abstract}
   Humans involuntarily tend to infer parts of the conversation from lip movements when the speech is absent or corrupted by external noise. In this work, we explore the task of lip to speech synthesis, i.e., learning to generate natural speech given only the lip movements of a speaker. Acknowledging the importance of contextual and speaker-specific cues for accurate lip-reading, we take a different path from existing works. We focus on learning accurate lip sequences to speech mappings for individual speakers in unconstrained, large vocabulary settings. To this end, we collect and release a large-scale benchmark dataset, the first of its kind, specifically to train and evaluate the single-speaker lip to speech task in natural settings. We propose a novel approach with key design choices to achieve accurate, natural lip to speech synthesis in such unconstrained scenarios for the first time. Extensive evaluation using quantitative, qualitative metrics and human evaluation shows that our method is four times more intelligible than previous works in this space. Please check out our demo video for a quick overview of the paper, method, and qualitative results. \url{https://www.youtube.com/watch?v=HziA-jmlk_4&feature=youtu.be}
\end{abstract}

\section{Introduction}
\label{section:intro}
Babies actively observe the lip movements of people when they start learning to speak~\cite{lewkowicz2012infants}. As adults, we pay high attention to lip movements to ``visually hear'' the speech in highly noisy environments. Facial actions, specifically the lip movements, thus reveal a useful amount of speech information. This fact is also exploited by individuals hard of hearing, who often learn to lip read their close acquaintances over time~\cite{iezzoni2004communicating} to engage in more fluid conversations. Naturally, the question arises as to whether a model can learn to voice the lip movements of a speaker by ``observing" him/her speak for an extended period. Learning such a model would only require videos of people talking with no further manual annotation. It also has a variety of practical applications such as (i) video conferencing in silent environments, (ii) high-quality speech recovery from background noise~\cite{afouras2018conversation}, (iii) long-range listening for surveillance and (iv) generating a voice for people who cannot produce voiced sounds (aphonia). Another interesting application would be ``voice inpainting"~\cite{zhou2019vision}, where the speech generated from lip movements can be used in place of a corrupted speech segment.

\begin{figure}[t]
  \includegraphics[width=\linewidth]{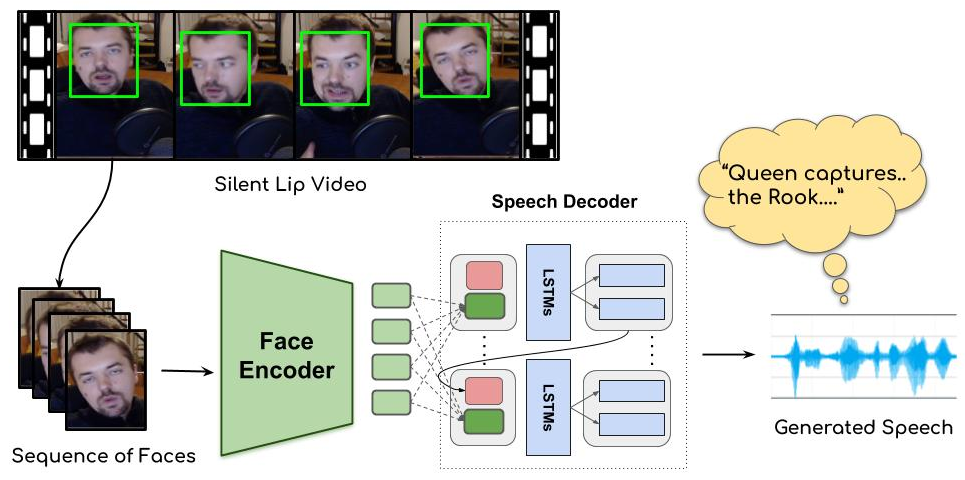}
  \caption{We propose ``\modelnamewithoutspace": a sequence-to-sequence architecture for accurate speech generation from silent lip videos in unconstrained settings for the first time. The text in the bubble is manually transcribed and is shown for presentation purposes.}
  \label{fig:firstpage}
\end{figure}

Inferring the speech solely from the lip movements, however, is a notoriously difficult task. A major challenge~\cite{ephrat2017vid2speech,chung2017lip} is the presence of homophenes: multiple sounds (phonemes) that are auditorily distinct being perceptually very similar with almost identical lip shapes (viseme). For instance, the lip shape when uttering the phoneme /p/ (\underline{pa}rk) can be easily confused with /b/ (\underline{ba}rk), and /m/ (\underline{ma}rk). As a matter of fact, only 25\% to 30\% of the English language is discernible through lip-reading alone~\cite{lieu2007communication,ebert1995communication,margellos2005developing,iezzoni2004communicating}. This implies that professional lip readers do not only lip-read but also piece together multiple streams of information: the familiarity with their subjects, the topic being spoken about, the facial expressions and head gestures of the subject and also their linguistic knowledge. In contrast to this fact, contemporary works in lip to speech and lip to text take a drastically different approach. 

Recent attempts in lip to text~\cite{Afouras2018DeepLR,chung2017lip} learn from unconstrained, large vocabulary datasets with thousands of speakers. However, these datasets only contain about $2$ minutes of data per speaker which is insufficient for models to learn concrete speaker-specific contextual cues essential for lip-reading. The efforts in lip to speech also suffer from a similar issue, but for a different reason. These works are constrained to small datasets~\cite{cooke2006audio} with narrow vocabulary speech in artificially constrained environments. 

In this work, we explore the problem of lip to speech synthesis from a unique perspective. We take inspiration from the fact that deaf individuals or professional lip readers find it easier to lip read people who they frequently interact with. Thus, rather than attempting lip to speech on random speakers in the wild, we focus on learning speech patterns of a specific speaker by simply observing the person's speech for an extended period. We explore the following question from a data-driven learning perspective: \textit{``How accurately can we infer an individual’s speech style and content from his/her lip movements?"}.

To this end, we collect and publicly release a $120$-hour video dataset of $5$ speakers uttering natural speech in unconstrained settings. Our \modelname dataset contains $800\times$ more data per speaker than the current multi-speaker datasets~\cite{Afouras2018DeepLR} to facilitate accurate modeling of speaker-specific audio-visual cues. The natural speech is spread across a diverse vocabulary\footnote{only words with frequency $> 4$ are considered} that is about $100\times$ larger than the current single speaker lip to speech datasets~\cite{cooke2006audio,harte2015tcd}. To the best of our knowledge, our dataset is the only publicly available large-scale benchmark to evaluate single-speaker lip to speech synthesis in unconstrained settings. With the help of this dataset, we develop \modelnamewithoutspace, a sequence-to-sequence model to generate accurate, natural speech that matches the lip movements of a given speaker. We support our results through extensive quantitative and qualitative evaluations and ablation studies. Our key contributions are as follows:

\begin{itemize}
    \item We investigate the problem of silent lip videos to speech generation in large vocabulary, unconstrained settings for the first time.
    \item We release a novel 120-hour person-specific \modelname dataset specifically for learning accurate lip to speech models of individual speakers. Each speaker contains $80\times$ more data with $100\times$ larger vocabulary than its counterparts. The speech is uttered in natural settings with no restriction to head pose or sentence lengths.
    \item Our sequence-to-sequence modeling approach produces speech that is almost $4\times$ more intelligible in unconstrained environments compared to the previous works. Human evaluation studies also show that our generated speech is more natural with rich prosody.
\end{itemize}

We release the data\footnote{\url{http://cvit.iiit.ac.in/research/projects/cvit-projects/speaking-by-observing-lip-movements}}, code\footnote{\url{https://github.com/Rudrabha/Lip2Wav}}, trained models publicly for future research along with a demonstration video here\footnote{\url{https://www.youtube.com/watch?v=HziA-jmlk_4}}. The rest of the paper is organized as follows: In Section \ref{section:relatedwork}, we
survey the recent developments in this space. Following this, we describe our novel \modelname dataset in Section \ref{section:dataset}. Our approach and training details are explained in Sections \ref{section:approach} and \ref{section:training_details}. We evaluate and compare our model with previous works in Section \ref{section:experiments}. We perform various ablation studies in Section \ref{section:ablation} and conclude our work in Section \ref{section:conclusion}.

\section{Related Work}
\label{section:relatedwork}

\begin{figure*}[t]
  \includegraphics[width=\textwidth]{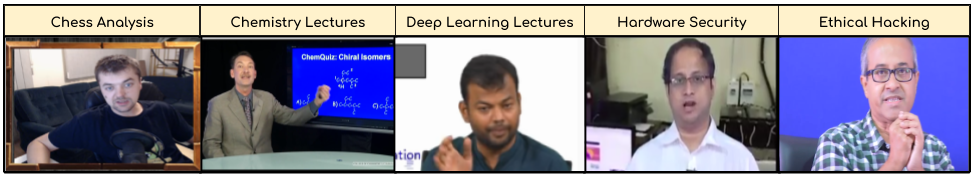}  \caption{Our \modelname dataset contains talking face videos of $5$ speakers from chess analysis and lecture videos. Each speaker has about $20$ hours of YouTube video content spanning a rich vocabulary of $5000+$ words.}
  \label{fig:dataset}
\end{figure*}

\subsection{Lip to Speech Generation}
While initial approaches~\cite{le2017generating,kello2004neural} to this problem extracted the visual features from sensors or active appearance models, the recent works employ end-to-end approaches. Vid2Speech~\cite{ephrat2017vid2speech} and Lipper~\cite{kumar2019lipper} generate low-dimensional LPC (Linear Predictive Coding) features given a short sequence of $K$ frames $(K < 15)$. The facial frames are concatenated channel-wise and a 2D-CNN is used to generate the LPC features. We show that this architecture is very inadequate to model real-world talking face videos that contain significant head motion, silences and large vocabularies. Further, the low dimensional LPC features used in these works do not contain a significant amount of speech information leading to robotic, artificial sounding speech. 

A follow-up work~\cite{Ephrat2017ImprovedSR} of Vid2Speech does away with LPC features and uses high-dimensional melspectrograms along with optical flows to force the network to explicitly condition on lip motion. While this can be effective for the GRID corpus that has no head movements, optical flow could be a detrimental feature in unconstrained settings due to large head pose changes. Another work~\cite{vougioukas2019video} strives for improved speech quality by generating raw waveforms using GANs. However, both these works do not make use of the well-studied sequence-to-sequence paradigm~\cite{sutskever2014sequence} that is used for text-to-speech generation~\cite{shen2018natural}; thus leaving a large room for improvement in speech quality and correctness. 

Finally, all the above works show results primarily on the GRID corpus~\cite{cooke2006audio} which has a very narrow vocabulary of $56$ tokens and very minimal head motion. We are the first to study this problem in a large vocabulary setting with thousands of words and sentences. Our datasets are collected from YouTube video clips and hence contain a significant amount of natural speech variations and head movements. This makes our results directly relevant to several real-world applications.

\subsection{Lip to Text Generation}
In this space as well, several works~\cite{chung2016lip,xu2018lcanet,wand2016lipreading,qu2019lipsound} are limited to narrow vocabularies and small datasets, however, unlike lip to speech, there have been multiple works~\cite{chung2017lip,Afouras2018DeepLR} that specifically tackle open vocabulary lip to text in-the-wild. They employ transformer sequence-to-sequence~\cite{vaswani2017attention} models to generate sentences given a silent lip movement sequence. These works also highlight multiple issues in the space of lip-reading, particularly the inherent ambiguity and hence the importance of using a language model. Our task at hand is arguably harder, as we not only have to infer the linguistic content, but also generate with rich prosody in the voice of the target speaker. Thus, we focus on extensively analyzing the problem in a \textit{single-speaker} unconstrained setting, and learning precise individual speaking styles. 

\subsection{Text to Speech Generation}
Over the recent years, neural text-to-speech models~\cite{shen2018natural,ping2017deep} have paved the way to generate high-quality natural speech conditioned on any given text. Using sequence-to-sequence learning~\cite{sutskever2014sequence} with attention mechanisms, they generate melspectrograms in an auto-regressive manner. In our work, we propose \modelnamewithoutspace, a modified version of Tacotron 2~\cite{shen2018natural} that conditions on face sequences instead of text.

\section{Speaker-specific \modelname Dataset}
\label{section:dataset}

The current datasets for lip to speech (or) text are at the two opposite ends of the spectrum: $(i)$ small, constrained narrow vocabulary like GRID~\cite{cooke2006audio}, TCD-TIMIT~\cite{harte2015tcd} or $(ii)$ unconstrained, open vocabulary multi-speaker like LRS2~\cite{Afouras2018DeepLR}, LRW~\cite{chung2016lip} and LRS3~\cite{afouras2018lrs3}. The latter class of datasets contains only about $2$ - $5$ minutes of data per speaker, making it significantly harder for models to learn speaker-specific visual cues that are essential for inferring accurate speech from lip movements. Further, the results would also be directly affected by the existing challenges of multi-speaker speech synthesis~\cite{gibiansky2017deep,jia2018transfer}. In the other extreme, the single-speaker lip to speech datasets~\cite{cooke2006audio,harte2015tcd}, do not emulate the natural settings as they are constrained to narrow vocabularies and artificial environments. Thus, both of these extreme cases do not test the limits of unconstrained single-speaker lip to speech synthesis. 

We introduce a new benchmark dataset for unconstrained lip to speech synthesis that is tailored towards exploring the following line of thought: \textit{How accurately can we infer an individual’s speech style and content from his/her lip movements?} To create the \modelname dataset, we collect a total of about $120$ hours of talking face videos across $5$ speakers. The speakers are from various online lecture series and chess analysis videos. We choose English as the sole language of the dataset. With about $20$ hours of natural speech per speaker and vocabulary sizes over $5000$ words\footnote{approximate; texts obtained using Google ASR API} for each of them, our dataset is significantly more unconstrained and realistic than GRID~\cite{cooke2006audio} or TIMIT~\cite{harte2015tcd} datasets. It is thus ideal for learning and evaluating accurate person-specific models for the lip to speech task. Table \ref{tab:dataset} compares the features of our \modelname dataset with other standard single-speaker lip-reading datasets. Note that a word is included in the vocabulary calculation for Table \ref{tab:dataset} only if its frequency in the dataset is at least five.

\begin{table}[ht]
    \centering
    {\footnotesize
    \begin{tabularx}{\linewidth}{|c|Y|Y|Y|Y|Y|}
    \hline
         \multirow{3}{1.5cm}{\centering Dataset} & Num. speakers & Total \#hours & \#hours per speaker & Vocab per speaker & Natural setting? \\ \hline
        GRID~\cite{cooke2006audio} & 34 & 28 & 0.8 & 56 & $\times$\\
        TIMIT~\cite{harte2015tcd} & 3 & 1.5 & 0.5 & 82 & $\times$ \\
        ~\textbf{\modelname (Ours)} & 5 & \textbf{120} & \textbf{$\approx$ 20} & \textbf{$\approx$ 5K} & \textbf{\checkmark}\\
    \hline
    \end{tabularx}
    }
    \caption{The \modelname dataset is the first large-scale dataset tailored towards acting as a reliable benchmark for single-speaker lip to speech synthesis.}
    \label{tab:dataset}
\end{table}

\section{Lip to Speech Synthesis in the Wild}
\label{section:approach}

Given a sequence of face images $I = (I_1, I_2, \dots, I_T)$ with lip motion, our goal is to generate the corresponding speech segment $S = (S_1, S_2, \dots, S_{T^{'}})$. To obtain natural speech in unconstrained settings, we make numerous key design choices in our \modelname architecture. Below, we highlight and discuss how they are different from previous methods for lip to speech synthesis.

\subsection{Problem Formulation}
Prior works in lip to speech regard their speech representation as a 2D-image~\cite{ephrat2017vid2speech,vougioukas2019video} in the case of melspectrograms or as a single feature vector~\cite{ephrat2017vid2speech} in the case of LPC features. In both these cases, they use a 2D-CNN to decode these speech representations. By doing so, they violate the ordering in which they model the sequential speech data, i.e. the future time steps influence the prediction of the current time step. In contrast, we formulate this problem in the standard sequence-to-sequence learning paradigm~\cite{sutskever2014sequence}. Concretely, each output speech time-step $S_k$ is modelled as a conditional distribution of the previous speech time-steps $S_{< k}$ and the input face image sequence $I = (I_1, I_2, \dots, I_T)$. The probability distribution of each output speech time-step is given by:

\begin{equation}
{P(S | I) = \Pi_{k} (S_k | S_{< k}, I)}
\end{equation}

\begin{figure}[h]
  \includegraphics[width=\linewidth]{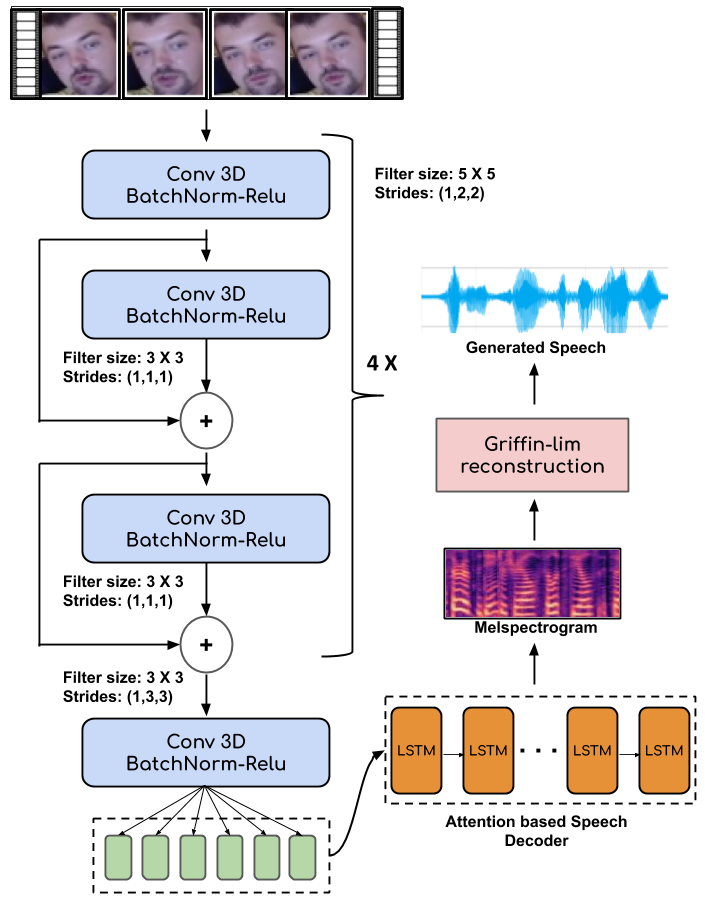}\caption{\modelname model for lip to speech synthesis. The spatio-temporal encoder is a stack of 3D convolutions to extract the sequence of lip movements. This is followed by a decoder adapted from~\cite{shen2018natural} for high-quality speech generation. The decoder is conditioned on the face image features from the encoder and generates the melspectrogram in an auto-regressive fashion.}
  \label{fig:architecture}
\end{figure}

\modelnamewithoutspace, as shown in Figure \ref{fig:architecture} consists of two modules: (i) Spatio-temporal face encoder (ii) Attention-based speech decoder. The modules are trained jointly in an end-to-end fashion. The sequence-to-sequence approach enables the model to learn an implicit speech-level language model that helps it to disambiguate homophenes. 

\subsection{Speech Representation} There are multiple output representations from which we can recover intelligible speech, but each of them have their trade-offs. The LPC features are low-dimensional and easier to generate, however, they result in robotic, artificial sounding speech. At the other extreme~\cite{vougioukas2019video}, one can generate raw waveforms but the high dimensionality of the output (16000 samples per second) makes the network training process computationally inefficient. We take inspiration from previous text-to-speech works~\cite{ping2017deep,shen2018natural} and generate melspectrograms conditioned on lip movements. We sample the raw audio at $16$kHz. The window-size, hop-size and mel dimension are $800, 200,$ and $80$ respectively.

\subsection{Spatio-temporal Face Encoder}
Our visual input is a short video sequence of face images. The model must learn to extract and process the fine-grained sequence of lip movements. 3D convolutional neural networks have been shown to be effective~\cite{ji20123d,tran2015learning,vougioukas2019video} in multiple tasks involving spatio-temporal video data. In this work, we try to encode the spatio-temporal information of the lip movements using a stack of 3D convolutions (Figure \ref{fig:architecture}). The input to our network is a sequence of facial images of the dimension $T\times H\times W\times 3$, where $T$ is the number of time-steps (frames) in the input video sequence, $H, W$ correspond to the spatial dimensions of the face image. We gradually down-sample the spatial extent of the feature maps and preserve the temporal dimension $T$. We also employ residual skip connections~\cite{he2016deep} and batch normalization~\cite{ioffe2015batch} throughout the network. The encoder outputs a single $D$-dimensional vector for each of the $T$ input facial images to get a set of spatio-temporal features $T \times D$ to be passed to the speech decoder. Each time-step of the embedding generated from the encoder also contains information about the future lip movements and hence helps in the subsequent generation.

\subsection{Attention-based Speech Decoder}
To achieve high-quality speech generation, we exploit the recent breakthroughs~\cite{shen2018natural,ping2017deep} in text-to-speech generation. We adapt the Tacotron 2~\cite{shen2018natural} decoder which has been used to generate melspectrograms conditioned on text inputs. For our work, we condition the decoder on the encoded face embeddings from the previous module. We refer the reader to the Tacotron 2~\cite{shen2018natural} paper for more details about the decoder. The encoder and decoder are trained end-to-end by minimizing the L1 reconstruction loss between the generated and the ground-truth melspectrogram. 

\subsection{Gradual Teacher Forcing Decay}
In the initial stages of training, up to $\approx30K$ iterations, we employ teacher forcing similar to the text-to-speech counterpart. We hypothesize that this enables the decoder to learn an implicit speech-level language model to help in disambiguating homophenes. Similar observations are made in lip to text works~\cite{Afouras2018DeepLR} which employ a transformer-based sequence-to-sequence model. Over the course of the training, we gradually decay the teacher forcing to enforce the model to attend to the lip region and to prevent the implicit language model from over-fitting to the train set vocabulary. We examine the effect of this decay in sub-section \ref{subsection:teacherforcing}.

\subsection{Context Window Size}
The size of the visual context window for inferring the current speech time-step helps the model to disambiguate homophenes~\cite{ephrat2017vid2speech}. We employ about $6\times$ larger context size than prior works and show in sub-section \ref{subsection:contextwindow} that this design choice results in significantly more accurate speech.


\section{Benchmark Datasets and Training Details}
\label{section:training_details}
\subsection{Datasets}
The primary focus of our work is on single-speaker lip to speech synthesis in unconstrained, large vocabulary settings. For the sake of comparison with previous works, we also train the \modelname model on the GRID corpus~\cite{cooke2006audio} and the TCD-TIMIT lip speaker corpus~\cite{harte2015tcd}. Next, we train on all five speakers of our newly collected speaker-specific \modelname dataset. Unless specified, all the datasets are divided into 90-5-5\% train, validation and unseen test splits. In the \modelname dataset, we create these splits using different videos ensuring that no part of the same video is used for both training and testing. The train and test splits are also released for fair comparison in future works.

\subsection{Training Methodology and Hyper-parameters}
\label{subsection:hparams}
We prepare a training input example by randomly sampling a contiguous sequence of $3$ seconds which corresponds to $T=75$ or $T=90$ depending on the frame rate (FPS) of the video. The effect of various context window sizes is studied in Section \ref{subsection:contextwindow}. We detect and crop the face from the video frames using the $S^3FD$ face detector~\cite{zhang2017s3fd}. The face crops are resized to $48 \times 48$. The melspectrogram representation of the audio corresponding to the chosen short video segment is used as the desired ground-truth for training. For training on small datasets like GRID and TIMIT, we halve the hidden dimension to prevent over-fitting. We set the training batch size to $32$ and train until the mel reconstruction loss plateaus for at least $30K$ iterations. In our experiments for unconstrained single-speaker, convergence was achieved in about $200K$ iterations. The optimizer used is Adam~\cite{kingma2014adam} with an initial learning rate of $10^{-3}$. The model with the best performance on the validation set is chosen for testing and evaluation. More details, specifically a few minor speaker-specific hyper-parameter changes can be found in the publicly released code\footnotemark[2].

\subsection{Speech Generation at Test Time}
During inference, we provide only the sequence of lip movements and generate the speech in an auto-regressive fashion. Note that we can generate speech for any length of lip sequences. We simply take consecutive $T$ second windows and generate the speech for each of them independently and concatenate them together. We also maintain a small overlap across the sliding windows to adjust for boundary effects. We obtain the waveform from the generated melspectrogram using the standard Griffin-Lim algorithm~\cite{Griffin1983SignalEF}. We observed that neural vocoders~\cite{Oord2016WaveNetAG} perform poorly in our case as our generated melspectrograms are significantly less accurate than state-of-the-art TTS systems. Finally, the ability to generate speech for lip sequences of any length is worth highlighting as the performance of the current lip-to-text works trained at sentence-level deteriorates sharply for long sentences that barely last over just $4$ - $5$ seconds~\cite{Afouras2018DeepLR}. 

\section{Experiments and Results}
\label{section:experiments}
We obtain results from our \modelname model on all the test splits as described above. For comparing related work, we use the open-source implementations provided by the authors if available or re-implement a version ourselves. We compare our models with the previous lip to speech works using three standard speech quality metrics: Short-Time Objective Intelligibility (STOI)~\cite{stoi} and Extended Short-Time Objective Intelligibility (ESTOI)~\cite{estoi} for estimating the intelligibility and Perceptual Evaluation of Speech Quality (PESQ)~\cite{pesq} to measure the quality. Using an out-of-the-box ASR system\footnote{Google Speech-to-Text API}, we obtain textual transcripts for our generated speech and evaluate our speech results using word error rates (WER) for the GRID~\cite{cooke2006audio} and TCD-TIMIT lip speaker corpus~\cite{harte2015tcd}. We, however do not compute WER for the proposed \modelname corpus due to the lack of text transcripts. We also perform human evaluation and report the mean opinion scores (MOS) for the proposed \modelname model and the competing methods. Next, we also perform extensive ablation studies for our approach and report our observations. Finally, as we achieve superior results compared to previous works in single-speaker settings, we end the experimental section by also reporting baseline results for word-level multi-speaker lip to speech generation using the LRW~\cite{chung2016lip} dataset and highlight its challenges as well.

\subsection{Lip to Speech in Constrained Settings}
We start by evaluating our approach against previous lip to speech works in constrained datasets, namely the GRID~\cite{cooke2006audio} corpus and TCD-TIMIT lip speaker corpus~\cite{harte2015tcd}. For the GRID dataset, we report the mean test scores for $4$ speakers which are also reported in the previous works. Tables \ref{tab:grid} and \ref{tab:timit} summarize the results for GRID and TIMIT datasets respectively. 

\begin{table}[h]
\setlength{\tabcolsep}{5pt}
\centering
  \begin{tabular}{|l|ccc|c|}
    \hline
    Method & STOI & ESTOI & PESQ & WER\\
    \hline
    Vid2Speech~\cite{ephrat2017vid2speech} & 0.491 & 0.335 & 1.734 & 44.92\%\\
    Lip2AudSpec~\cite{Akbari2017Lip2AudspecSR} & 0.513 & 0.352 & 1.673 & 32.51\%\\
    GAN-based~\cite{vougioukas2019video} & 0.564 & 0.361 & 1.684 & 26.64\%\\
    Ephrat et al.~\cite{Ephrat2017ImprovedSR} & 0.659 & 0.376 & \textbf{1.825} & 27.83\%\\
    \textbf{\modelname (ours)} & \textbf{0.731} & \textbf{0.535} & 1.772 & \textbf{14.08\%}\\
  \hline
\end{tabular}
    \vspace{0.2cm}
    \caption{Objective speech quality, intelligibility and WER scores for the GRID dataset unseen test split.}
    \label{tab:grid}
\end{table}

\begin{table}[h]
\setlength{\tabcolsep}{5pt}
\centering
  \begin{tabular}{|l|ccc|c|}
    \hline
    Method & STOI & ESTOI & PESQ & WER\\
    \hline
    Vid2Speech~\cite{ephrat2017vid2speech} & 0.451 & 0.298 & 1.136 & 75.52\%\\
    Lip2AudSpec~\cite{Akbari2017Lip2AudspecSR} & 0.450 & 0.316 & 1.254 & 61.86\%\\
    GAN-based~\cite{vougioukas2019video} & 0.511 & 0.321 & 1.218 & 49.13\%\\
    Ephrat et al.~\cite{Ephrat2017ImprovedSR} & 0.487 & 0.310 & 1.231 & 53.52\%\\
    \textbf{\modelname (ours)} & \textbf{0.558} & \textbf{0.365} & \textbf{1.350} & \textbf{31.26\%}\\
  \hline
\end{tabular}
    \vspace{0.2cm}
    \caption{Objective speech quality, intelligibility and WER scores for the TCD-TIMIT dataset unseen test split.}
    \label{tab:timit}
\end{table}

As we can see, our approach outperforms competing methods across all objective metrics by a significant margin. The difference is particularly visible in the TIMIT~\cite{harte2015tcd} dataset, where the test set contains a lot of novel words unseen during training. This shows that our model learns to capture correlations across short phoneme sequences and pronounces new words better than previous methods.
\subsection{Lip to Speech in Unconstrained Settings}
We now move on to evaluating our approach in unconstrained datasets that contain a lot of head movements and much broader vocabularies. They also contain a significant amount of silences or pauses between words and sentences. It is here that we see a more vivid difference in our approach compared to previous approaches. We train our model independently on all $5$ speakers of our newly collected \modelname dataset. The training details are mentioned in the sub-section \ref{subsection:hparams}. For comparison with previous works, we choose the best performing models~\cite{vougioukas2019video, Ephrat2017ImprovedSR} on the TIMIT dataset based on STOI scores and report their performance after training on our \modelname dataset. We compute the same metrics for speech intelligibility and quality that are used in Table \ref{tab:timit}. The scores for all five speakers for our method and the two competing methods across all three metrics are reported in Table \ref{tab:lip2speechscores}.

\begin{table}
\setlength{\tabcolsep}{5pt}
\centering
  \begin{tabular}{|l|c|ccc|}
    \hline
    Method & Speaker & STOI & ESTOI & PESQ \\
    \hline
    GAN-based~\cite{vougioukas2019video} & \multirow{3}{1.5cm}{\centering\textit{Chemistry Lectures}} & 0.192 & 0.132 & 1.057\\
    Ephrat et al.~\cite{Ephrat2017ImprovedSR}&& 0.165 & 0.087 & 1.056 \\
    \textbf{\modelname (ours)}&& \textbf{0.416} & \textbf{0.284} & \textbf{1.300} \\
    \hline
    GAN-based~\cite{vougioukas2019video} & \multirow{3}{1.5cm}{\centering\textit{Chess Analysis}} & 0.195 & 0.104 & 1.165 \\
    Ephrat et al.~\cite{Ephrat2017ImprovedSR}&& 0.184 & 0.098 & 1.139\\
    \textbf{\modelname (ours)}&& \textbf{0.418} & \textbf{0.290} & \textbf{1.400}\\
    \hline
    GAN-based~\cite{vougioukas2019video} & \multirow{3}{1.5cm}{\centering\textit{Deep Learning}} & 0.144 & 0.070 & 1.121 \\
    Ephrat et al.~\cite{Ephrat2017ImprovedSR}&& 0.112 & 0.043 & 1.095 \\
    \textbf{\modelname (ours)}&& \textbf{0.282} & \textbf{0.183} & \textbf{1.671} \\
    \hline
    GAN-based~\cite{vougioukas2019video} & \multirow{3}{1.5cm}{\centering\textit{Hardware Security}} & 0.251 & 0.110 & 1.035\\
    Ephrat et al.~\cite{Ephrat2017ImprovedSR}&& 0.192 & 0.064 & 1.043 \\
    \textbf{\modelname (ours)}&& \textbf{0.446} & \textbf{0.311} & \textbf{1.290} \\
    \hline
    GAN-based~\cite{vougioukas2019video} & \multirow{3}{1.5cm}{\centering\textit{Ethical hacking}} & 0.171 & 0.089 & 1.079\\
    Ephrat et al.~\cite{Ephrat2017ImprovedSR}&& 0.143 & 0.064 & 1.065 \\
    \textbf{\modelname (ours)}&& \textbf{0.369} & \textbf{0.220} & \textbf{1.367} \\
  \hline
 \end{tabular}
    \vspace{0.2cm}
    \caption{In unconstrained single-speaker settings, our \modelname model achieves almost $4\times$ more intelligible speech than the previous methods.}
    \label{tab:lip2speechscores}
\end{table}

Our approach produces much more intelligible and natural speech across different speakers and vocabulary sizes. Notably, our model has more accurate pronunciation, as can be seen in the increased STOI and ESTOI scores compared to the previous works.

\subsection{Human Evaluation}
In addition to speech quality and intelligibility metrics, it is important to manually evaluate the speech as these metrics are not perfect~\cite{Ephrat2017ImprovedSR} measures.

\subsubsection{Objective Human Evaluation}
In this study, we ask the human participants to manually identify and report (A) the percentage of mispronunciations, (B) the percentage of word skips and (C) the percentage of mispronunciations that are homophenes. Word skips denotes the number of words that are either completely unintelligible due to noise or slurry speech. We choose $10$ predictions from the unseen test split of each speaker in our \modelname dataset to get a total of $50$ files. We report the mean numbers of (A), (B), and (C) in Table \ref{tab:objectivehumaneval}.

\begin{table}[h]
\centering
  \begin{tabular}{|c|ccc|}
    \hline
    Model & (A)  & (B) & (C) \\
    \hline
    GAN-based~\cite{vougioukas2019video} & 36.6\% & 24.3\% & 63.8\%\\
    Ephrat et al~\cite{Ephrat2017ImprovedSR} & 43.3\% & 27.5\% & 60.7\%\\
    \textbf{\modelname (ours)}  & \textbf{21.5\%} & \textbf{8.6\%} & \textbf{49.8\%} \\
  \hline
\end{tabular}
    \vspace{.2cm}
    \caption{Objective Human evaluation results. The participants manually identified the percentage of (A) Mispronunciations, (B) Word skips and (C) Homophene-based errors in the test samples.}
    \label{tab:objectivehumaneval}
\end{table}

Our approach makes far fewer mispronunciations than the current state-of-the-art method. It also skips words $3\times$ lesser, however, the key point to note is that the issue of homophenes is still a dominant cause for errors in all cases indicating there is still scope for improvement in this area.

\subsubsection{Subjective Human Evaluation}
We ask $15$ participants to rate the different approaches for unconstrained lip to speech synthesis on a scale of $1 - 5$ for each of the following criteria: (i) \textit{Intelligibility} and (ii) \textit{Naturalness} of the generated speech. Using $10$ samples of generated speech for each of the $5$ speakers from our \modelname dataset, we compare the following approaches: (i) Our \modelname model (ii) Current state-of-the-art lip to speech models~\cite{vougioukas2019video,Ephrat2017ImprovedSR} (iii) Manually transcribed text followed by a multi-speaker TTS~\cite{shen2018natural,jia2018transfer} to show that even with the most accurate text, lip to speech is not a concatenation of lip-to-text and text-to-speech. And finally, (iv) Human speech is also added for reference. In all the cases, we overlay the speech on the face video before showing it to the rater. The mean scores are reported in Table \ref{tab:subjectivehumaneval}.

\begin{table}[h]
\centering
  \begin{tabular}{|c||c|c|}
    \hline
    Approach & Intelligibility & Naturalness \\
    \hline
    GAN-based~\cite{vougioukas2019video} & 1.56 & 1.71\\
    Ephrat et al.~\cite{Ephrat2017ImprovedSR} & 1.34 & 1.67\\
    \textbf{\modelname (ours)} & \textbf{3.04} & \textbf{3.63}\\
    \hline
    MTT + TTS~\cite{shen2018natural} & 3.86 & 3.15\\
    Actual Human Speech & 4.82 & 4.95 \\
  \hline
\end{tabular}
    \vspace{.2cm}
    \caption{Mean human evaluation scores based on speech quality and intelligibility for various approaches for lip to speech. MTT denotes ``manually-transcribed text". The penultimate row simulates the best possible case of automatic lip to text followed by a state of the art text-to-speech system. The drop in naturalness score in this case illustrates the loss in speech style and prosody.}
    \label{tab:subjectivehumaneval}
\end{table}

In line with the previous evaluations, we can see that our approach produces significantly higher quality and legible speech compared to the previous state-of-the-art~\cite{vougioukas2019video}. It is also evident that generating the speech from the text that is read from lip movements (lip to text), cannot achieve the desired prosody and naturalness even if the text is fully accurate. Further, this method will also cause the lips and audio to be out of sync. Thus, our approach is currently the best method to produce natural speech from lip movements. 

\subsection{Multi-speaker Word-level Lip to Speech}
Given the superior performance of our \modelname approach on single-speaker lip to speech, we also obtain baseline results on the highly challenging problem of multi-speaker lip to speech synthesis for random identities. Note that the focus of the work is still primarily on single-speaker lip to speech. We adapt the approach presented in~\cite{jia2018transfer} and feed a speaker embedding as input to our model. We report our baseline results on the LRW~\cite{chung2016lip} dataset intended for word-level lip-reading, i.e. it is used to measure the performance of recognizing a single word in a given short phrase of speech. We do not demonstrate on the LRS2 dataset~\cite{chung2017lip} as its clean train set contains just $29$ hours of data, which is quite small for multi-speaker speech generation. For instance, multi-speaker text-to-speech generation datasets~\cite{zen2019libritts} containing a similar number of speakers contain several hundreds of hours of speech data.

In Table \ref{tab:multispeakerscores}, we report the speech quality and intelligibility metrics achieved by our multi-speaker \modelname model on the LRW test split. As none of the previous works in lip to speech tackle the multi-speaker case, we do not make any comparisons with them. We also report the WER by getting the text using the Google ASR API. For comparison, we also report the WER of the baseline lip to text work on LRW~\cite{chung2016lip}. Note that the speech metric scores shown in Table \ref{tab:multispeakerscores} for word-level lip to speech cannot be directly compared with the single-speaker case which contains word sequences of various lengths along with pauses and silences.

\begin{table}[h]
\centering
  \begin{tabular}{|l|ccc|c|}
    \hline
    Method & STOI & ESTOI & PESQ & WER\\
    \hline
    \modelname (Ours) & 0.543 & 0.344 & 1.197 & ~\textbf{34.2\%} \\
    Chung et al.~\cite{chung2016lip} & NA & NA & NA & 38.8\%\\
  \hline
\end{tabular}
    \vspace{.2cm}
    \caption{Objective speech quality and intelligibility scores on the LRW dataset. WER is also calculated after using an ASR on the generated speech. Our model outperforms the baseline method proposed in~\cite{chung2016lip}, without any text-level supervision. The speech metrics are not applicable for~\cite{chung2016lip} as it is a lip to text work.}
    \label{tab:multispeakerscores}
\end{table}

We end our experimental section here. Apart from showing significant increases in performance from previous lip to speech works, we also achieve word-level multi-speaker lip to speech synthesis. In the next section, we conduct ablation studies on our model.
\section{Ablation Studies}
\label{section:ablation}
In this section, we probe different aspects of our \modelname approach. All results in this section are calculated using the unseen test predictions on the ``Hardware Security" speaker of our \modelname dataset. 

\subsection{Larger context window helps in disambiguation}
\label{subsection:contextwindow}
As stated before, the lip to speech task is highly ambiguous to be inferred solely from lip movements. One of the ways to combat this, is to provide reasonably large context information to the model to disambiguate a given viseme. Previous works, however, use only about $0.3 - 0.5$ seconds of context. In this work, we use close to $6\times$ this number and provide a context of $3$ seconds. This helps the model to disambiguate by learning co-occurrences of phonemes and words and the resulting improvement is evident in Table \ref{tab:contextwindows}. 
\begin{table}[h]
\centering
  \begin{tabular}{|c||c|c|c|}
    \hline
    Context Window size & STOI & ESTOI & PESQ \\
    \hline
    0.5 seconds & 0.264 & 0.193 & 1.062\\
    1.5 seconds & 0.321 & 0.226 & 1.080\\
    \textbf{3 seconds} &\textbf{0.446} & \textbf{0.311} & \textbf{1.290}\\
  \hline
\end{tabular}
    \vspace{.2cm}
    \caption{Larger context information consistently results in more accurate speech generation. We limit the window size to $3$ seconds due to memory constraints.}
    \label{tab:contextwindows}
\end{table}

\subsection{Model is Highly Attentive to the Mouth}

\begin{figure}[h]
  \includegraphics[width=\linewidth]{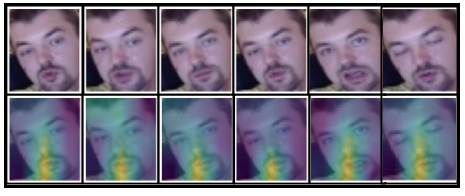}
  \caption{We plot the activations of the penultimate layer of the face encoder and the attention alignment from the decoder. We see that the face encoder is highly attentive towards the mouth region.}
  \label{fig:heatmap}
\end{figure}
\begin{figure}[h]
  \includegraphics[width=\linewidth]{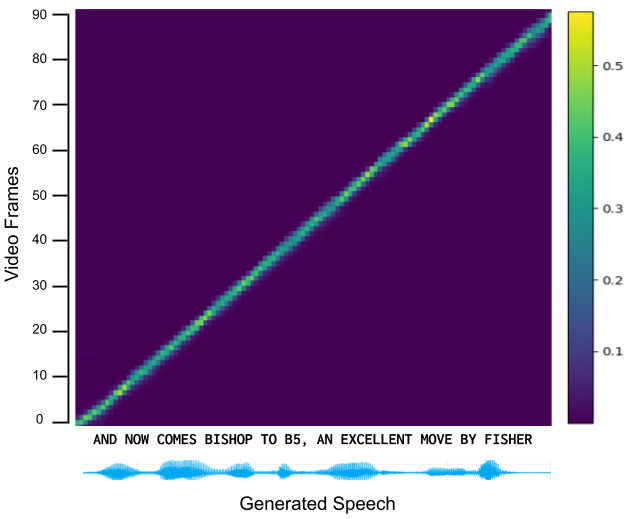}
  \caption{The decoder alignment curve illustrates that the model is generating speech by strongly conditioning on the corresponding lip movements.}
  \label{fig:alignment}
\end{figure}

We plot the activations of the penultimate layer of the spatio-temporal face encoder in Figure \ref{fig:heatmap} to show that our encoder is highly attentive towards the mouth region of the speaker. The attention alignment curve in Figure \ref{fig:alignment} shows that the decoder conditions on the appropriate video frame's lips while generating the corresponding speech.

\subsection{Teacher Forcing vs Non-Teacher Forcing}
\label{subsection:teacherforcing}
To accelerate the training of a sequence-to-sequence architecture, typically, the previous time step's ground-truth (instead of the generated output) is given as input to the current time-step. While this is highly beneficial in the initial stages of training, we observed that gradually decaying the teacher forcing from $\approx 30K$ iterations significantly improves results and prevents over-fitting to the train vocabulary. A similar improvement is also observed in lip to text works~\cite{Afouras2018DeepLR}. In Table \ref{tab:teacherforcing}, we show the significant improvement in test scores by gradually decaying teacher forcing.

\begin{table}[h]
\centering
  \begin{tabular}{|c||c|c|c|}
    \hline
    Teacher-forcing & STOI & ESTOI & PESQ \\
    \hline
    Always forced & 0.221 & 0.162 & 1.141\\
    Gradual decay & \textbf{0.446} & \textbf{0.311} & \textbf{1.290} \\
  \hline
\end{tabular}
    \vspace{.2cm}
    \caption{Gradually decaying the teacher forcing enables the model to generalize to unseen vocabulary by forcing it to look at the visual input and not just predict from the previously uttered speech.}
    \label{tab:teacherforcing}
\end{table}

\subsection{Effect of Different Visual Encoders}
While using a 3D-CNN worked best in our experiments to capture both the spatial and temporal information in unconstrained settings, we also report in Table \ref{tab:diffencoders} the effect of using different kinds of encoders. We replace the encoder module while keeping the speech decoder module intact. We see that the best performance is obtained with a 3D-CNN encoder.

\begin{table}[h]
\centering
 \begin{tabular}{|c||c|c|c|}
    \hline
    Encoder & STOI & ESTOI & PESQ \\
    \hline
    2D-CNN & 0.291 & 0.211 & 1.112\\
    2D-CNN + 1D-CNN & 0.298 & 0.223 & 1.170 \\
    \textbf{3D-CNN (ours)} &\textbf{0.446} & \textbf{0.311} & \textbf{1.290}\\
  \hline
\end{tabular}
    \vspace{.2cm}
    \caption{Our \modelname model employs a 3D-CNN encoder to capture the spatio-temporal visual information and is the superior choice over the other alternatives.}
    \label{tab:diffencoders}
\end{table}

\section{Conclusion}
\label{section:conclusion}
In this work, we investigated the problem of synthesizing speech based on lip movements. We specifically solved the problem by focusing on individual speakers. We did this in a data-driven learning approach by creating a large-scale benchmark dataset for unconstrained, large vocabulary single-speaker lip to speech synthesis. We formulate the task at hand as a sequence-to-sequence problem, and show that by doing so, we achieve significantly more accurate and natural speech than previous methods. We evaluate our model with extensive quantitative metrics and human studies. All the code and data for our work has been made publicly available\footnotemark[2]. Our work opens up several new directions. One of them would be to examine related works in this space such as lip to text generation from a speaker-specific perspective. Similarly, explicitly addressing the dominant issue of homophenes can yield more accurate speech. Generalizing to vocabulary outside the typical domain of the speaker can be another fruitful venture. We believe that exploring some of the above problems in a data-driven fashion could lead to further useful insights in this space.

{\small
\bibliographystyle{ieee_fullname}
\bibliography{references}
}

\end{document}